\documentclass[runningheads]{llncs}
\usepackage[T1]{fontenc}
\usepackage{graphicx,verbatim}
\usepackage{multirow}
\usepackage{amsmath, amssymb}
\usepackage{amsmath, amssymb}
\usepackage{booktabs}
\begin{document}
\title{Conformal forecasting for surgical instrument trajectory }
\author{Sara Sangalli\inst{1} \and Gary Sarwin\inst{1} \and Ertunc Erdil\inst{1} \and Alessandro Carretta\inst{2,3}  \and \\  Victor Staartjes\inst{2}  \and
Carlo Serra\inst{2} \and
Ender Konukoglu\inst{1}}
\authorrunning{Sangalli et al.}
\institute{Computer Vision Lab, ETH Zurich, Zurich, Switzerland \and
Department of Neurosurgery, University Hospital of Zurich, Zurich, Switzerland \and
Department of Biomedical and Neuromotor Sciences, University of Bologna, Bologna, Italy}

\maketitle              
\begin{abstract}

Forecasting surgical instrument trajectories and predicting the next surgical action recently started to attract attention from the research community. 
Both these tasks are crucial for automation and assistance in endoscopy surgery. 
Given the safety-critical nature of these tasks, reliable uncertainty quantification is essential. 
Conformal prediction is a fast-growing and widely recognized framework for uncertainty estimation in machine learning and computer vision, offering distribution-free, theoretically valid prediction intervals. 
In this work, we explore the application of standard conformal prediction and conformalized quantile regression to estimate uncertainty in forecasting surgical instrument motion, i.e., predicting direction and magnitude of surgical instruments' future motion. 
We analyze and compare their coverage and interval sizes, assessing the impact of multiple hypothesis testing and correction methods. Additionally, we show how these techniques can be employed to produce useful uncertainty heatmaps.
To the best of our knowledge, this is the first study applying conformal prediction to surgical guidance, marking an initial step toward constructing principled prediction intervals with formal coverage guarantees in this domain. 

\end{abstract}

\section{Introduction}
Surgical guidance is a key challenge in computer-assisted interventions, aiming to support surgeons during procedures. 
In neurosurgery, neuronavigation systems are commonly used to enhance spatial orientation, providing clinicians with valuable real-time information during procedures \cite{hartl2013worldwide,orringer2012neuronavigation}. 
Recent advances in machine learning have opened opportunities to obtain a comprehensive understanding of the surgical scene and provide guidance in the form of surgical phase recognition, anatomy detection, workflow recognition and more \cite{garrow2021machine,das2024pitvis,padoy2019machine,sarwin2024vision,staartjes2021machine}. 
In addition, very recent efforts have been made to provide surgical guidance through surgical instrument trajectory forecasting \cite{sarwin2025anatomyneedforecastingsurgery}.

This type of prediction has multiple applications, primarily advancing surgical assistance beyond self-orientation by anticipating the next movement and providing real-time guidance.  
Furthermore, forecasting as proposed in \cite{sarwin2025anatomyneedforecastingsurgery} can enable unsupervised extraction of best surgical practices from expert surgeons, simply by leveraging large video databases maintained by several institutions. These algorithms could enhance training and guidance for less experienced surgeons and potentially serve as a foundation for autonomous surgery.

However, accurately forecasting surgical instrument trajectories is inherently challenging, as it requires predicting future motion. Due to temporal extrapolation, this process introduces an additional layer of uncertainty, 
which must be carefully accounted for when deploying machine learning-based forecasting algorithms \cite{unc_surg_navig_1,unc_surg_navig_2}. Ensuring safe usage, as well as fostering trust and acceptance among clinicians \cite{tschandl2020human}, requires mechanisms to quantify and visualize this uncertainty. A valuable, and arguably essential, tool in this regard would be an \textit{uncertainty map}, for instance in the form of a heatmap. Such a visualization could highlight potential future locations of the surgical instrument, while providing information about the uncertainty in these regions. 
Beyond assisting surgeons, uncertainty maps can guide automated surgical systems in risk assessment, e.g. choosing the best motion within an area with a maximum user-specified uncertainty level.


To ensure that such uncertainty maps are reliable and theoretically solid, uncertainty quantification through Conformal Prediction (CP) \cite{cp_gentle_intro,cp_regression,distr_free_pr_sets} can be employed. Crucially, CP can provide distribution-free, model-agnostic and theoretically valid \textit{predictive intervals} (PIs) for the predicted trajectories. These intervals are guaranteed to contain the true outcome with a user-specified probability $1-\alpha$ under mild assumptions.
When applied to surgical trajectory forecasting, CP generates uncertainty maps that can build trust in automated systems and support real-time decision-making by guaranteeing \textit{desired confidence levels} for predicted actions.
Specifically, the most popular variant, Split CP \cite{cp_regression}, ensures \textit{statistical validity} by leveraging the predictions on a calibration set to determine PIs for new test samples. Alternative methods for uncertainty calibration or quantification, such as temperature scaling \cite{on_cal_nn}, Monte Carlo Dropout \cite{mcdrop} or Deep Ensembles \cite{deepens}, can adjust the softmax probabilities produced by neural networks to provide pseudo-confidence levels, however, they lack robust guarantees that conformal prediction can provide.

While CP has been increasingly explored in medical imaging applications \cite{multiview_cp_miccai,conforma_volume_estimation_miccai,unc_cp_miccai}, to the best of our knowledge, its application in the surgical setting remains unexplored. Integrating CP-based uncertainty quantification could be an important step toward developing more reliable and interpretable deep learning models for real-time decision-making during surgery.

\textbf{Contributions.} 
In this paper, we apply conformal techniques to the challenging task of forecasting surgical instrument trajectories in endoscopic video sequences. Building on \cite{sarwin2025anatomyneedforecastingsurgery}, which predicts the next instrument position relative to the last observed frame, we extend this framework by separately quantifying uncertainty in both the angle and magnitude of the predicted motion vector, ensuring that the ground truth statistically falls within the PIs with a user-specified probability $1-\alpha$.
We show that independently constructed PIs fail to maintain valid joint coverage, as expected from hypothesis testing theory \cite{vovk_book}, and address this with multiple-testing corrections. We systematically compare split CP and conformalized quantile regression \cite{cqr} on a pituitary surgery dataset, analyzing coverage and PI widths. Additionally, we demonstrate how these PIs enable uncertainty heatmaps with statistical guarantees (Figure \ref{fig:examples_intervals}). 
Finally, we discuss the challenges of maintaining valid coverage in real-world surgical data and highlight the potential of CP in surgical forecasting, highlighting its potential in surgical forecasting and encouraging further exploration in this direction.
\section{Methods}
\subsection{Problem definition}
In this work we build our task and models consistently with \cite{sarwin2025anatomyneedforecastingsurgery}. Essentially, this framework processes a sequence of endoscopic frames, which we denote as $s_t = x_{t-d:t} := \{x_\tau\}_{\tau=t-d}^t$ with $x_\tau$ representing a frame at time $\tau$ and $d$ is the sequence length. 
These sequences are processed by an object detection model which detects anatomical structures and the surgical instrument for every frame in $s_t$. The detections are then fed into a model that consists of an encoder which maps them to a latent representation $z_{t}$. 
Subsequently, this latent representation is processed by a decoder that predicts the changes relative to frame $t$ in the bounding box locations and sizes of the surgical instrument for the next $h$ frames.
Each bounding box is represented by four coordinates, namely the center coordinates of the box and the corresponding height and width. 

In this work we focus on the center coordinates of the instrument bounding box and discard the changes in box width and height. 
The changes in the $h$ future bounding box centers are modeled by a single overall movement vector, which is the vector sum of relative changes between successive frames. 
We refer to it as $v$ for the ground truth (GT) vector, and as $\hat{v}$ the predicted vector.
Effectively, $v$ represents the change in bounding box center between frame $t$ and $t+h$, and $\hat{v}$ its prediction. 
The vector $v$, as well as its prediction, can be described with its phase $\angle v$, measured relative to the positive $x$-axis, and magnitude $||v||$. This separation of direction and scale provides a more interpretable representation for analyzing the predicted motion. We aim to build uncertainty intervals for $\angle \hat{v}$ and $||\hat{v}||$, which are guaranteed to contain the ground truth $\angle{v}$ and $||v||$ with at least a user specified probability $1-\alpha$, with the aid of conformal procedures. 

\subsection{Background on Conformal Prediction}
We consider the standard split conformal prediction (CP) setting \cite{cp_gentle_intro}, where a dataset is split to obtain a hold-out calibration set $D_{cal} = \{(f_i, y_i)\}_{i=1}^{n}$ in a problem where $f$ represents input features and $y$ the prediction.  Given a new test point $f_{n+1}$ we seek to predict $y_{n+1}$ while assuming $D_{cal} \cup (f_{n+1}, y_{n+1})$ is exchangeable, i.e., its joint distribution is invariant under permutation, implying identically distributed but not necessarily independent samples. The goal is to construct a prediction interval $PI_\alpha$ that contains $y_{n+1}$ with at least probability of $1-\alpha$, where $\alpha \in (0,1)$ is a user-specified error rate.

\textbf{Procedure.} We first define a non-negative \textit{conformity score} $R(f,y)$ for each sample in $D_{cal}$, quantifying the discrepancy between the predicted value $\hat{y}_i$ and the ground truth $y_i$. \\ We compute $Q_{1-\alpha}=(1 - \alpha)\left(1 + \frac{1}{| D_{cal}|}\right)\text{-th empirical quantile of } \{R_i : i \in D_{cal}\}$. The PI for a test sample is then: $PI_\alpha = [\hat{y}_{n+1}-Q_{1-\alpha}, \hat{y}_{n+1}+Q_{1-\alpha}]$. This interval is guaranteed to satisfy the desired \textit{marginal coverage} $1-\alpha$ \cite{vovk_book}, i.e.
\begin{equation}\label{eq:coverage}
    \mathbb{P} \left(  y_{n+1} \in PI_{\alpha}(f_{n+1}) \right) \geq 1 - \alpha.
\end{equation}
The \textit{marginal coverage} represents the proportion over the test dataset for which the true outcome falls within the PI, ensuring a user-specified confidence level, e.g., 70\%. We refer to it simply as \textit{coverage}.

\begin{figure}[ht] 
    \centering
    \includegraphics[width=0.8\textwidth]{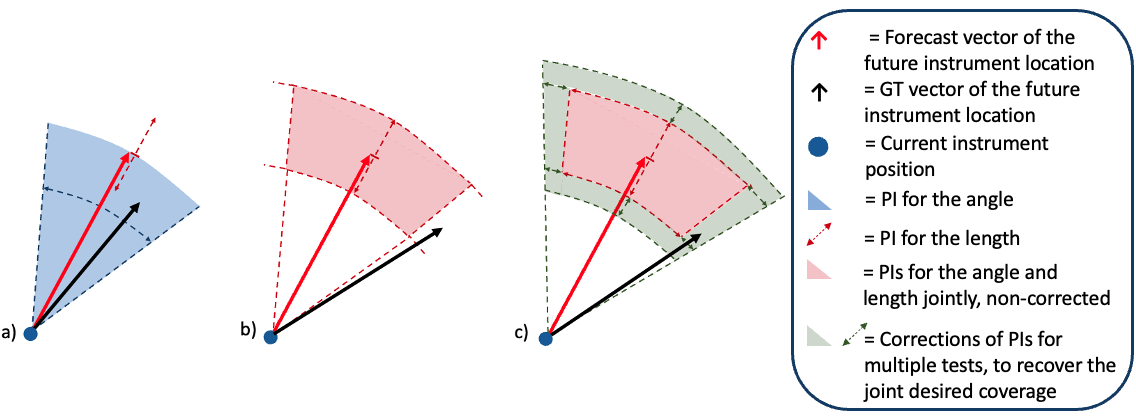} 
    \caption{Qualitative illustration of how Conformal Prediction (CP) for instrument trajectory forecasting. \textit{a)}: CP applied independently to the angle and the length. \textit{b)}: Joint intervals obtained by merging the independent ones without corrections, here failing to cover the angle. \textit{c)}: Multiple-test corrections restore valid coverage for both quantities.}
    \label{fig:method_explanation} 
\end{figure}

In our task of instrument trajectory forecasting, we define $PI_{\alpha}$ as a range of angles or magnitudes to contain the phase or magnitude of the GT vector $v$ with a user-specified coverage level $1-\alpha$. We further define a joint interval that considers both the phase and the magnitude at the same time, taking into account multiple testing corrections. 
Figure~\ref{fig:method_explanation}.a illustrates qualitative PIs for both magnitude and phase, treated independently. In the next Subsection we describe the algorithms more formally for our setup, focusing on the angle prediction; the same algorithms also hold for the magnitude prediction.

\subsection{Conformal strategies for surgical tool trajectory forecasting}

\textbf{Split Conformal Prediction.} Let us consider an exchangeable dataset $D_{cal}$, composed of a set of video sequences $s_1, ..., s_n$, their corresponding ground truth motion vectors $v_1, ..., v_n$ and the corresponding predicted vectors $\hat{v}_1, ..., \hat{v}_n$. Let $s_{n+1}$ be a new test example drawn from the same distribution as the samples in $D_{cal}$ . 
For a regression problem as ours, split CP \cite{cp_gentle_intro} computes the conformity scores as the absolute error residuals, for instance for the angle $R^{CP}_i = |\angle v_i- \angle \hat{v}_i|, i\in D_{cal}$.
For a desired coverage $1-\alpha$, we compute the quantile of the empirical distribution of the absolute residuals
\[
Q_{1-\alpha}(R^{CP}, D_{cal}) := (1 - \alpha)\left(1 + \frac{1}{| D_{cal}|}\right)\text{-th empirical quantile of } \{R^{CP}_i : i \in D_{cal}\}.
\]
Then, the prediction interval for the angle for the new sample $s_{n+1}$ is given by
\[    PI_{\alpha}(s_{n+1}) = \left[\angle \hat{v}_{n+1} - Q_{1-\alpha}(R^{CP}, D_{cal}), \angle \hat{v}_{n+1} + Q_{1-\alpha}(R^{CP}, D_{cal})\right]
,\]
where $\angle \hat{v}_{n+1}$ is the predicted angle and the $PI$ satisfies the marginal coverage guarantee     $\mathbb{P} \left( \angle v_{n+1} \in PI_{\alpha}(s_{n+1}) \right) \geq 1 - \alpha$.

\textbf{Conformalised Quantile Regression (CQR).} To produce more adaptive intervals for the desired coverage $1-\alpha$, we also explore how CQR \cite{cqr} performs in our setting. First, CQR employs regression networks $\hat{Q}(s)$ that need to be trained on a training set $D_{train}$ with a pinball loss \cite{cqr}, to produce lower and upper quantile predictions denoted with $\{\hat{q}_{\frac{\alpha}{2}}(s), \hat{q}_{1- {\frac{\alpha}{2}}}(s)\}$, respectively. The pinball loss is defined as:
\begin{equation}
\mathcal{L}_\alpha(y, \hat{y}) := 
\begin{cases} 
\alpha(y - \hat{y}) & \text{if } y - \hat{y} > 0, \\ 
(1 - \alpha)(\hat{y} - y) & \text{otherwise}
\end{cases}\label{eq:pinball}
\end{equation}
Then, CQR computes the conformity scores that quantify the error made by the two regressed quantiles as  $R^{CQR}_i = max\{\hat{q}_{\frac{\alpha}{2}}(s_i) - \angle v_i, \angle v_i - \hat{q}_{1- {\frac{\alpha}{2}}}(s_i)\}, i\in D_{cal}$. Finally, given a new input $s_{n+1}$, we construct the prediction interval for $v_{n+1}$ as 
\[
    PI_{\alpha}(s_{n+1}) = \left[\hat{q}_{\frac{\alpha}{2}} (s_{n+1}) - Q_{1-\alpha}(R^{CQR}, D_{cal}), \hat{q}_{1- \frac{\alpha}{2}} (s_{n+1}) + Q_{1-\alpha}(R^{CQR}, D_{cal})\right].
\]
This interval also satisfies $\mathbb{P} \left( \angle v_{n+1} \in PI_{\alpha}(s_{n+1}) \right) \geq 1 - \alpha$ while providing sample-adaptive PIs.

\subsection{Multiple testing corrections}
So far, we have discussed how to apply conformal procedures to a single quantity at a time, namely the phase (or similarly the magnitude) of the forecast vector in our setting. A natural extension is to construct a joint predictive interval that simultaneously accounts for both quantities. A straightforward approach to obtaining such a joint interval is to compute the intersection of the individually obtained PIs. 
However, this approach introduces multiple testing issues, as the probability of at least one quantity falling outside the joint interval increases, reducing the overall coverage guarantee \cite{vovk_book,two_steps_cp_det}. In our application, it would mean that only one of the two PIs—either for phase or magnitude— would successfully cover the GT vector. An example is qualitatively presented in Figure \ref{fig:method_explanation}.b. As a consequence, this results in an effective coverage guarantee of at most $(1-k\cdot\alpha)$, with number of tests $k=2$ in our case, rather than the desired $(1-\alpha)$. We refer to \cite{two_steps_cp_det} for a mathematical proof.

Various correction techniques have been proposed in the literature on individual test level to restore valid coverage guarantees \cite{simes,sidak,max_rank,Vovk2012CombiningPV}. Figure \ref{fig:method_explanation}.c provides a qualitative example of how the joint coverage could be restored thanks to quantiles corrections, for  a case where originally only the magnitude of the vector was successfully falling in the PI. The most commonly used Bonferroni correction \cite{Vovk2012CombiningPV} proposes to select a new $\alpha_{corr} = \alpha / k $ on individual tests, in order to effectively guarantee a global coverage $1-\alpha$. We also explore a less conservative variant of it, called Sidak correction \cite{sidak}, that sets $\alpha_{corr} = 1-(1-\alpha)^{1/k}$, that assumes independence across tests.
Finally, we explored the Max-Rank correction, which operates directly on the ranking of nonconformity scores across all variables. It ensures that the worst-case (i.e., maximum) ranked score across all tests is used for constructing the prediction interval. To apply it to our setup, we scaled the conformity scores of angles and length to the range $[0,1]$ to make them comparable and finally scaled them back to their original range.
\section{Experiments}
\textbf{Dataset.} The dataset consists of 144 pituitary surgery videos (transsphenoidal adenomectomy) from unique patients, collected over a 10‐year period across multiple centers using various endoscopes under general research consent. Expert neurosurgeons annotated 16 classes—15 anatomical structures (typically one per video) and 1 surgical instrument class (comprising multiple instruments). 77 videos (approx. 10,000 labeled images) were used to train and validate a YOLO \cite{yolo} object detection backbone. The trained object detection model was used to detect the anatomy and instruments in each frame of 57 videos, which then served to train and validate a surgical action forecasting model. The remaining 10 videos were reserved for testing; all test frames were processed by the detection model to form the final test set. The 57 videos training set is also used to fine-tune the CQR heads and the validation set for CQR model selection.
The test set is further split into 6 patients for calibrating the conformal procedures and 4 patients for evaluation on unseen patients.
As the calibration dataset is randomly sampled and predictions are made independently for each sequence without memory, the exchangeability assumption required for conformal predictions holds. 

\noindent\textbf{Forecasting network.} We adopt the architecture from \cite{sarwin2025anatomyneedforecastingsurgery}, consisting of a transformer encoder and three fully connected layers, producing a 16-dimensional latent representation $z_{t}$. A single linear layer serves as the decoder.  Given 64 input frames, the model predicts the next 8-frame trajectory, with a regression error of 47° for angle and 0.2 for length, normalized to the image dimensions \textit{(h, w}).

\noindent\textbf{CQR network.}  Like in \cite{two_steps_cp_det}, we keep the pre-trained encoder frozen and train four quantile regression heads for CQR, to predict the quantiles $\{\hat{q}_{\frac{\alpha}{2}}(s), \hat{q}_{1- {\frac{\alpha}{2}}}\}$, for both the vector angle and magnitude.
The architecture consists of four fully connected layers with ReLU activations, batch normalization, and dropout regularization. The output layer produces predictions for the two quantiles across two distinct regression tasks (angle and magnitude), corresponding to the two conformal tests applied. To enforce meaningful predictions, the angular outputs are mapped to the range $[-\pi, \pi]$ using a tanh transformation, while the magnitude outputs are constrained to $[0, \sqrt{2}]$ using a sigmoid activation. The two new decoders are trained with a pinball loss \cite{cqr} for 500 epochs.

\begin{figure}[t] 
    \centering
    \includegraphics[width=0.8\textwidth]{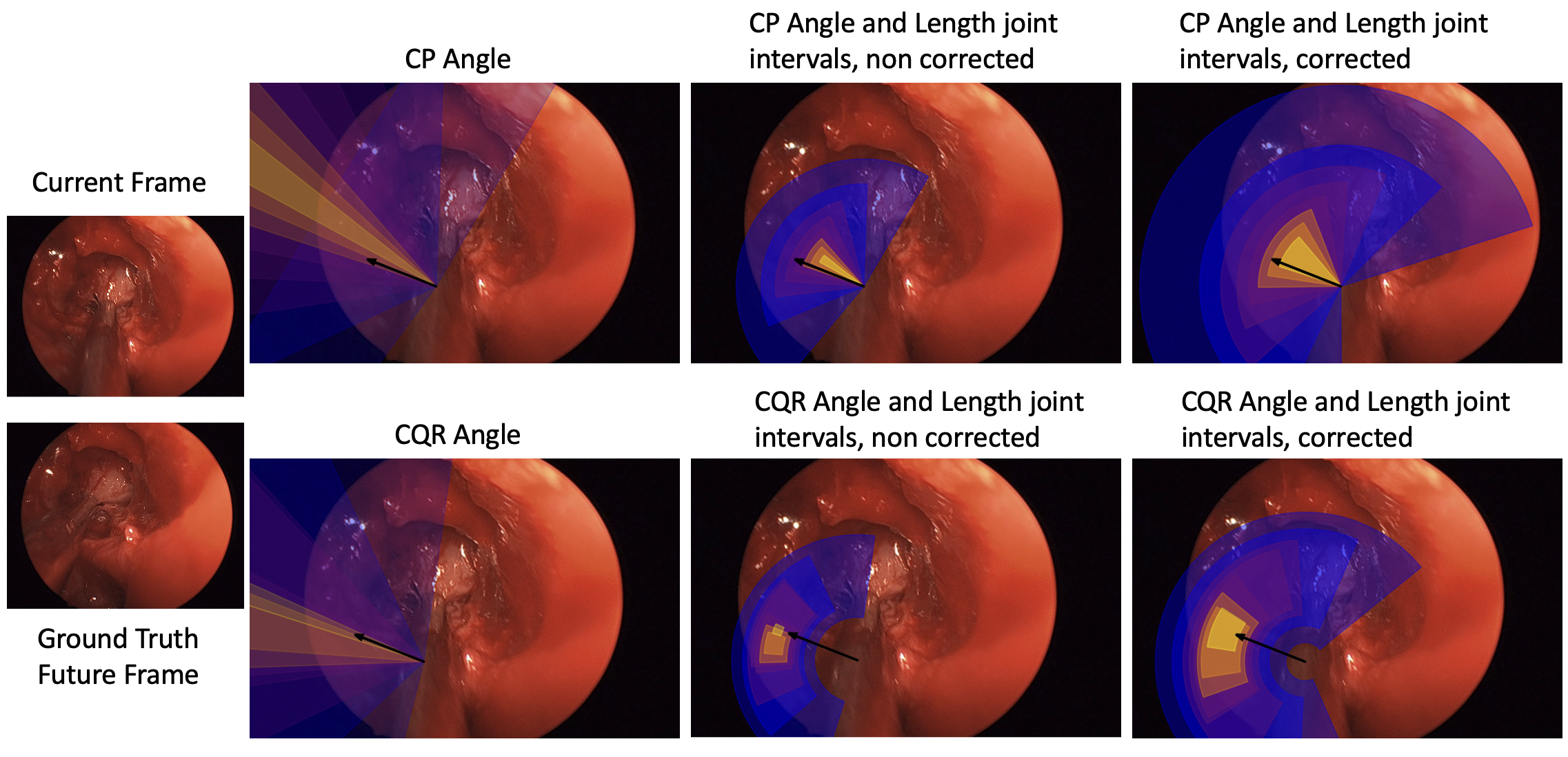} 
    \caption{Heatmaps from CP (top) and CQR (bottom). The black vector denotes the GT trajectory. Target coverage ranges from 10\% (yellow) to 80\% (blue).  \textit{Left}: Angle-only intervals—CQR yields sharper intervals and better coverage than CP.  \textit{Center}: Joint intervals without correction—coverage fails as expected. \textit{Right}: Sidak-corrected joint intervals—recalibration restores validity, with CQR providing tighter bounds.}
    \label{fig:examples_intervals} 
\end{figure}


\noindent\textbf{Computational overhead.} CP is a post-hoc method, adding minimal computational overhead as it only requires threshold computation on the validation set. CQR is similarly efficient, integrating lightweight regression heads with minimal impact on inference speed. Given these properties, conformal techniques are well-suited for real-time surgical guidance, providing rigorous uncertainty quantification without compromising inference speed.

\noindent\textbf{Protocol and Metrics.} We assess CP and CQR for angle and magnitude predictions independently. Then, we first demonstrate that directly combining these independent PIs fails to guarantee the desired joint coverage. To address this, we apply Bonferroni (Bonf.), Sidak, and Max-Rank corrections to restore valid coverage. Performance is evaluated in terms of empirical coverage, which should align with the target level, and PI width, which should be minimized while maintaining valid coverage. To account for variability of (marginal) coverage, each experiment is repeated 20 times with shuffled calibration and test sets, and we report the mean and standard deviation of the results.


\section{Results}
We report results for target coverages of 60\% and 70\% in Table \ref{tab:results}.  In general, CQR generates more precise PIs compared to CP. This is due to the quantile networks' ability to adaptively learn the data distribution for the specified intervals. In contrast, CP utilizes fixed thresholds derived from the calibration set, independent of the input. Noticeably, conformalization of length alone shows higher variability, especially for CP, likely due to the inherent noise in real-world surgical datasets. This variability arises when instruments are removed from the scene, introducing fluctuations in vector magnitude.  For joint intervals, as expected from theory, coverage drops significantly (by 25-30\%) without multiple-test corrections. Applying corrections successfully restores coverage, particularly for CQR, while naturally increasing PI sizes to ensure joint validity.

\begin{table}[th]
    \centering
    \caption{Empirical coverage and PI sizes for target coverage of 60\% and 70\%. CQR generally yields smaller PIs for single tests (bold) and most of corrected joint tests. As expected, uncorrected joint coverage drops significantly but is restored with corrections.}
    \label{tab:results}
    \begin{tabular}{c | c c | c c}
        \toprule
        \multirow{2}{*}{\textbf{Method}} & \multicolumn{2}{|c}{\textbf{Target Coverage = 60\%}} & \multicolumn{2}{|c}{\textbf{Target Coverage = 70\%}} \\
        \cmidrule(lr){2-3} \cmidrule(lr){4-5}
        & \textbf{Coverage (\%)} & \textbf{PI Size ($\downarrow)$} & \textbf{Coverage (\%)} & \textbf{PI Size ($\downarrow)$} \\
        \midrule
        CP angle & 59.7 $\pm$ 3.7 & 78.5°$\pm$ 3.6°& 69.8 $\pm$  2.3& 111.9° $\pm$ 3.4°\\
        CQR angle & 59.5$\pm$ 3.5 & \textbf{69.3° $\pm$ 3.1°} & 69.4 $\pm$ 2.8 & \textbf{103.5°  $\pm$ 3.4°} \\
        \cmidrule(lr){1-5} 
        CP length & 59.5 $\pm$ 12.1 &  0.25 $\pm$ 0.03  & 67.9 $\pm$ 12.9 & 0.31 $\pm$0.04\\
        CQR length & 60.4 $\pm$ 9.5 &\textbf{ 0.19 $\pm$ 0.02} & 69.2 $\pm$ 9.8 & \textbf{0.24 $\pm$ 0.03} \\
        \cmidrule(lr){1-5} 
        CP joint, non corr. & 31.0 $\pm$ 7.7 & $-$ & 43.8$\pm$ 9.0 & $-$ \\
        CQR joint, non corr. & 33.2 $\pm$ 5.4 & $-$ & 45.8 $\pm$ 6.9 & $-$ \\
        \cmidrule(lr){1-5} 
        CP joint,  Bonf. corr. & 59.3 $\pm$ 8.4 & 169.8°, 0.41  & 67.8 $\pm$ 6.4 & 211.6°, 0.5 \\
        CQR joint,  Bonf. corr. & 61.5 $\pm$ 6.6 & 165.9°, 0.17 & 70.0 $\pm$6.0 & 212.7°, 0.21\\
        CP joint,  Sidak corr. & 55.2 $\pm$ 8.7 & 151.8, 0.38 & 65.4 $\pm$ 7.7 & 200.0°, 0.46 \\
        CQR joint,  Sidak corr. & 57.2 $\pm$ 6.9 & 144.2°, 0.15 & 67.7 $\pm$ 6.2  & 200.5°,  0.19 \\
        CP joint,  Max-Rank corr. & 58.2 $\pm$ 7.9 & 105.7°, 0.83 & 68.3 $\pm$ 7.7 & 206.4°,  0.48\\
        CQR joint,  Max-Score corr. & 59.0  $\pm$ 4.2& 75.2°, 0.59 &   68.9 $\pm$ 4.0 & 204.8°,  0.20 \\
        \bottomrule
    \end{tabular}
\end{table}

\noindent\textbf{Uncertainty heatmaps with guarantees.} Figure \ref{fig:examples_intervals} presents heatmaps of the PIs at target coverages between 10\% and 80\%, that could be implemented in a surgical navigation system. CQR generally provides intervals that better centered around the GT vector, improving coverage for both the angle individually and for the joint predictions. For the joint intervals, CP provides too large length quantiles, that result in not having an annular sector, when compared to CQR. Finally, the corrected intervals, compared to the uncorrected ones, are more centered around the GT vector head, translating to higher coverage.

\section{Conclusion}
This work demonstrated the potential of conformal techniques for uncertainty quantification in surgical instrument trajectory forecasting. By modeling uncertainty in both angle and magnitude, we ensure PIs for the predicted trajectory with statistical guarantees. Our results show that CQR yields more precise PIs with lower variability than CP, and multiple-test corrections restore empirical joint coverage. We also illustrate how these intervals enable meaningful uncertainty maps for surgical guidance. Future work includes e.g. extending to autoregressive models, while addressing complexities like exchangeability violations.

\section{Acknowledgments}
This study was financially supported by: 1. The LOOP Z\"{u}rich – Medical Research Center, Zurich, Switzerland, 2. Personalized Health and Related Technologies (PHRT), project number 222, ETH domain and 3. Clinical Research Priority Program (CRPP) Grant on Artificial Intelligence in Oncological Imaging Network, University of Z\"{u}rich, 3. The SNSF (Project IZKSZ3\_218786).

We thank David Stutz for the insightful discussions on conformal prediction, which helped shape our understanding and approach in this work.

%
 \bibliographystyle{splncs04}
 \bibliography{mybibliography}

\end{document}